\documentclass{article}

\usepackage[numbers]{natbib}
\usepackage[preprint]{neurips_2024}

\usepackage[utf8]{inputenc} 
\usepackage[T1]{fontenc}    
\usepackage{hyperref}       
\usepackage{url}            
\usepackage{booktabs}       
\usepackage{amsfonts}       
\usepackage{nicefrac}       
\usepackage{microtype}      
\usepackage{xcolor}         

\usepackage{graphicx} 
\usepackage{amsmath}
\usepackage{amsfonts}
\usepackage{amssymb}
\usepackage{listings}
\usepackage{xcolor}
\usepackage{tikz}
\usetikzlibrary{3d,angles,quotes}
\usepackage{hyperref}
\usepackage{subcaption}
\usepackage{bm}
\usepackage{graphicx}%
\usepackage{multirow}%
\usepackage{amsmath,amssymb,amsfonts}%
\usepackage{amsthm}%
\usepackage{mathrsfs}%
\usepackage[title]{appendix}%
\usepackage{xcolor}%
\usepackage{textcomp}%
\usepackage{manyfoot}%
\usepackage{booktabs}%
\usepackage{algorithm}%
\usepackage{algorithmicx}%
\usepackage{algpseudocode}%
\usepackage{listings}%
\usepackage{hyperref}
\usepackage{Sty/mcr}
\usepackage{bm}
\usepackage{subcaption}

\usepackage{setspace}
\usepackage{times}
\usepackage{graphicx}
\usepackage{color}
\usepackage{multirow}
\usepackage{Sty/mcr}

\title{Crystal-LSBO: Automated Design of De Novo Crystals with Latent Space Bayesian Optimization}

\author{%
  Onur Boyar \\
  \small Nagoya University \\
  \small RIKEN \\
  \texttt{\tiny boyar.onur.nagoyaml@gmail.com} \\
  \And
  Yanheng Gu \\
  \small Nagoya University \\
  \texttt{\tiny gu.yanheng.nagoyaml@gmail.com} \\
  \And  
  Yuji Tanaka \\
  \small Nagoya University \\
  \texttt{\tiny tanaka.yuji.nagoyaml@gmail.com} \\
  \And
  Shunsuke Tonogai \\
  \small DENSO CORP. \\
  \texttt{\tiny shunsuke.tonogai.j3y@jp.denso.com} \\
  \And  
  Tomoya Itakura \\
  \small DENSO CORP. \\
  \texttt{\tiny tomoya.itakura.j8w@jp.denso.com} \\
  \And
  Ichiro Takeuchi\thanks{Corresponding author} \\
  \small Nagoya University \\
  \small RIKEN \\
  \texttt{\tiny takeuchi.ichiro.n6@f.mail.nagoya-u.ac.jp} \\
}

\begin{document}

\maketitle


\begin{abstract}
Generative modeling of crystal structures is significantly challenged by the complexity of input data, which constrains the ability of these models to explore and discover novel crystals. This complexity often confines de novo design methodologies to merely small perturbations of known crystals and hampers the effective application of advanced optimization techniques. One such optimization technique, Latent Space Bayesian Optimization (LSBO) has demonstrated promising results in uncovering novel objects across various domains, especially when combined with Variational Autoencoders (VAEs). Recognizing LSBO's potential and the critical need for innovative crystal discovery, we introduce Crystal-LSBO—a de novo design framework for crystals specifically tailored to enhance explorability within LSBO frameworks. Crystal-LSBO employs multiple VAEs, each dedicated to a distinct aspect of crystal structure—lattice, coordinates, and chemical elements, orchestrated by an integrative model that synthesizes these components into a cohesive output. This setup not only streamlines the learning process but also produces explorable latent spaces thanks to the decreased complexity of the learning task for each model, enabling LSBO approaches to operate. Our study pioneers the use of LSBO for de novo crystal design, demonstrating its efficacy through optimization tasks focused mainly on formation energy values. Our results highlight the effectiveness of our methodology, offering a new perspective for de novo crystal discovery.
\end{abstract}

\section{Introduction}

The discovery and design of novel crystals are crucial for advancements in a range of scientific and industrial fields. 
The ability to engineer crystals with specific, desired properties holds the potential to influence future technological developments significantly. 
Therefore, establishing a framework to automate the discovery of de novo crystals is vital for various industries. 
A promising approach to achieving this goal involves the application of machine learning, particularly through generative modeling.

The field of generative models for crystal design has emerged as a focal point for researchers from diverse disciplines. Recent literature highlights various efforts employing techniques such as diffusion models \cite{xie2021crystal, zeni2023mattergen}, generative adversarial networks (GANs) \cite{kim2020generative}, autoencoders (AEs) \cite{nohinverse}, and variational autoencoders (VAEs) \cite{ren2022invertible}. Each of these methods addresses the challenge of capturing the complex nature of crystals, which encompasses factors like lattice structures, types of elements, and their coordinates. This complexity contributes to an expansive search space for designing new crystals with desired properties. Consequently, a generative model capable of learning lower-dimensional representations of input instances proves invaluable. Such a model can significantly reduce the search space, enabling more efficient exploration for target crystals with specific features. Among the aforementioned methods, VAEs stand out for their ability to map complex crystal structures into a manageable, lower-dimensional \emph{latent space} and generate new instances from it, enabling efficient search space for de novo crystal designs. Thus, our study employs a VAE-based approach, leveraging VAEs' potential and their effective integration with Bayesian Optimization (BO) techniques, referred to as Latent Space Bayesian Optimization (LSBO)~\cite{gomez2018}. LSBO operates in the latent space of VAE to balance exploration and exploitation efficiently, with demonstrated success in the discovery of novel instances in diverse fields. However, the existing straightforward application of LSBO encounters difficulties in the exploration phase with the complexity of crystal structures, typically focusing only on exploitation, i.e., being able to generate valid outputs only within the immediate vicinity of latent representations of known crystals. Given the considerable time and costs involved in crystal design, there is an urgent need for a sample-efficient methodology that enables extensive exploration of de novo crystals.

To this end, we introduce Crystal-LSBO, a framework to design de novo crystals that integrates a specialized set of VAEs, explicitly tailored to facilitate LSBO methodologies. Crystal-LSBO is designed to simplify the learning of crystal structures by employing multiple VAEs, each responsible for distinct components of the crystal structure: lattices, coordinates, and elements. After each VAE captures the latent representations of different crystal components, these are then combined by an integrative VAE into a unified and comprehensive material latent space, which forms the search space for LSBO algorithms. This structured approach not only streamlines the learning process but also significantly improves our ability to navigate and explore latent spaces—crucial for the discovery of novel and optimal crystal designs via LSBO in a sample-efficient manner. 

To showcase the effectiveness of the Crystal-LSBO framework, we applied it to design de novo crystals with optimal formation energy values, a property that is related to the overall stability and functionality of materials. Results show the effectiveness of Crystal-LSBO, paving the way for new advancements in material science. The implementation of the Crystal-LSBO framework is available at \href{https://github.com/onurboyar/Crystal-LSBO}{https://github.com/onurboyar/Crystal-LSBO}. Our contributions are listed as follows:
\begin{enumerate}
    \item We introduce Crystal-LSBO, a unified framework using specialized VAE models for de novo discovery of crystals.
    \item We demonstrate the explorability of Crystal-LSBO's latent space, showcasing its ability to generate valid crystals across a broad region.
    \item We apply LSBO to crystal design through the Crystal-LSBO framework, demonstrating its effectiveness through experimental evaluations focused on formation energy. Our results are benchmarked against existing methodologies, showcasing significant improvements.
\end{enumerate}

\section{Related Works}\label{sec:related_works}

In crystal generative modeling, the intricate nature of crystal structure representations often requires employing complex generative models or using multiple models.
For instance, \citet{hoffmann2019data} employed a VAE paired with a U-Net to learn from 3D density maps, and \citet{court20203} proposed a Conditional-VAE and U-Net pairing.  
Another approach explored by \citet{nohinverse} utilizes two AEs and a VAE, where embeddings of lattice parameters and chemical elements are learned by separate AEs and used as input to the VAE model to form the generative model. 
Besides, the literature offers diverse methods for representing crystal structure data. 
\citet{chiba2022neural} introduced neural structure fields for use in AE-based architectures. 
\citet{ren2020inverse} developed an invertible representation combining real and reciprocal space features for VAE training. 
Building on this, \citet{ren2022invertible} refined this method by incorporating Fourier-transformed crystal properties (FTCP) into the VAE, referred to as FTCP-VAE. 
Although these studies employ encoding-based approaches, the high dimensionality of the latent spaces makes effective search and optimization challenging.
Diverging from VAEs, GANs also found themselves applications in crystal generation tasks with examples like \cite{kim2020generative, zhao2021high}. 
Besides, in \cite{zeni2023mattergen}, a diffusion model based crystal generation and optimization method is proposed, and \citet{xie2021crystal} employed a diffusion model in conjunction with a VAE. 
Although diffusion models demonstrate promising results in crystal design, their low sample efficiency due to high dimensional search space and the inherent stochasticity of the diffusion process hinder their effectiveness in de novo design tasks.

Despite variations in generative models and crystal representation techniques across the aforementioned studies, a common strategy for generating new crystals is to sample from the vicinity of known crystals in the latent space. This approach is primarily adopted for two reasons: first, the vast search space resulting from high-dimensional latent representations makes it challenging to explore effectively; second, sampling broadly across the latent space often leads to the generation of invalid crystal structures.


LSBO has garnered significant attention, particularly in the field of organic molecule design. It was first introduced in this domain through the seminal work by Gomez-Bombarelli et al.~\cite{gomez2018}. They represented organic molecules as sequences and developed a VAE to explore the latent space for de novo design. Subsequently, numerous studies have focused on method development and practical applications to improve and extend the LSBO framework~\cite{Grosnit2021HighDimensionalBO, tripp_etal_2020, maus2022local, griffiths_etal_2020, boyar2024latent}. The primary reason LSBO is particularly advanced in organic molecular design is that organic molecules can be represented as sequences using approaches such as SMILES~\cite{weininger_1988} and SELFIES~\cite{krenn_2020}. Unfortunately, for our target area of crystal design, there is no simple representation like SMILES, making it difficult to apply LSBO directly. To our knowledge, no existing studies have successfully applied LSBO to crystal design. In this study, we develop a method for LSBO in crystal design by constructing latent spaces for each component of a crystal, namely the lattice, coordinate, and element, and then developing a unified latent space that integrates these components, within which the LSBO is conducted. 

\section{Preliminaries and Problem Setup}\label{sec:preliminaries}

In this section, we provide preliminary knowledge on crystals, VAEs, and LSBO. We then discuss the challenges of property optimization and material space exploration using generative models.

\subsection{Crystals}

Crystals are characterized by an organized arrangement of elements, forming a three-dimensional lattice that repeats consistently throughout the material. This repeating structure is constructed from a basic unit called the unit cell, which acts as the fundamental building block of the crystal. The unique arrangement of elements within these unit cells dictates the material's physical and chemical properties.


Below are definitions of key crystal-related terms used in this paper:

\begin{itemize}
    \item \textbf{Lattice:} A lattice is a 3-dimensional framework that shows where atoms are located in a crystal. The structure of the lattice is defined by a set of vectors, called lattice vectors, along with their lengths and the angles between them, which together form the entire crystal structure. The lattice serves as the framework that organizes the atomic arrangement of the material.
    
    \item \textbf{Space Groups:} Space groups categorize the patterns and symmetry in crystal structures, formed by operations like translations, rotations, reflections, and inversions. There are 230 space groups, each defining unique symmetry operations and constraints about how the elements within the crystal and the lattice should be arranged  \cite{brock2016international}. Space group 1 is the simplest, involving only translations, meaning the pattern repeats periodically without additional symmetry. This group has minimal geometric constraints, allowing flexible lattice angles and lengths, and all crystals can be represented in this form by simplifying their original symmetry.

    \item \textbf{Primitive Cell:} The primitive cell is the smallest possible unit cell that, when repeated in the crystal space, fully describes the entire crystal structure. It contains only one lattice and represents the simplest building block of the crystal. The primitive cell is unique in that it has the minimum volume and includes all the symmetry and properties of the crystal lattice.
    
    \item \textbf{Coordinates:} Coordinates refer to the specific positions of atoms within the unit cell or primitive cell, typically expressed in terms of fractions of the lattice vectors. These fractional coordinates describe where each atom is located relative to the origin of the unit cell and are crucial for determining the exact structure of the crystal.
    
    \item \textbf{Elements within the Material:} The elements within the material refer to the specific types of atoms that make up the crystal or material structure. These elements are defined by their atomic types (such as carbon, oxygen, hydrogen, etc.) and are fundamental to the chemical composition of the material. The arrangement and interaction of these elements within the crystal lattice determine the material's physical and chemical properties. Figure \ref{fig:multi_vae}(A) demonstrates the components of the crystals.
\end{itemize}

\subsection{VAEs}\label{sec:2.1}
A VAE \cite{kingma_welling_2014} consists of an encoder $f_{\phi}^{\text{enc}}: \mathcal{X} \to \mathcal{Z}$ and a decoder $f_{\theta}^{\text{dec}}: \mathcal{Z} \to \mathcal{X}$, where $\mathcal{X}$ represents the input space and $\mathcal{Z}$ the latent space. The encoder maps an input $\bm{x}$ to a latent representation $\bm{z}$, while the decoder reconstructs the input from the latent space.

The encoder models the conditional probability $q_{\phi}(\bm{z} | \bm{x})$ as an approximation of the true posterior $p_{\theta}(\bm{z} | \bm{x})$, and the decoder models $p_{\theta}(\bm{x} | \bm{z})$. The VAE is trained by optimizing the following objective function:
\begin{equation}
\mathcal{L}^{\text{VAE}}(\theta, \phi; \bm{x}) = \mathbb{E}_{q_{\phi}(\bm{z} | \bm{x})} [\log p_{\theta}(\bm{x} | \bm{z})] - \beta \, D_{\text{KL}}(q_{\phi}(\bm{z} | \bm{x}) \| p(\bm{z})),
\end{equation}
where $D_{\text{KL}}$ denotes the Kullback-Leibler divergence, $\beta$ balances the reconstruction and regularization terms \cite{higgins2016}, and $p(\bm{z})$ is typically set as a standard multivariate normal distribution $\mathcal{N}(\bm{0}, \bm{I})$.


\subsection{LSBO}\label{sec:2.2}
In LSBO, we start with numerous \emph{unlabeled} instances $\{{\bm x_i}\}_{i \in [\cU]}$ and a smaller set of \emph{labeled} instances ${(\bm x_i, y_i)}_{i \in [\cL]}$, where $\bm x_i \in \cX$ represents inputs like crystal structures, and $y_i \in \cY \subseteq \RR$ are their labels, such as formation energy. The sets of indices of the unlabeled and labeled instances are denoted as $\cU$ and $\cL$, respectively. BO optimizes a costly black-box (BB) function $f^{\rm BB}: \cX \to \cY$. The goal is to maximize $f^{\rm BB}$ with minimal evaluations by employing a Gaussian Process (GP) as a surrogate model to predict the function’s behavior across $\cX$. Effective BO relies on a surrogate model to guide the selection of candidate inputs $\bm x$ that might yield values exceeding $\max_{i \in \cL} y_i$. However, creating an effective GP surrogate model in high-dimensional input spaces like those for crystal structures is challenging. LSBO addresses this by using a VAE to reduce dimensionality, training the VAE on the unlabeled set $\cU$ and fitting the GP model to the latent space $\cZ$. This approach simplifies the surrogate model fitting because $\cZ$ is of lower dimensionality than $\cX$. During LSBO iterations, the acquisition function (AF) applied to the GP model’s predictions selects new points in $\cZ$ to evaluate. The selected latent variable $\bm{z}_{i'} = \argmax_{\bm z \in \cZ} f^{\rm AF}(\bm z)$ is then decoded into a new input instance $\bm x_{i'} = f_\theta^{\rm dec}(\bm{z}_{i'})$. This new instance is evaluated by $f^{\rm BB}$, and the results update the labeled set $\cL$ and refine the GP model. Optionally, retraining the VAE with the updated data can integrate new findings into the model. This cycle repeats until achieving optimal results or exhausting resources. In contexts like de novo crystal design, LSBO aims to discover crystal structures with optimal properties, efficiently navigating the reduced dimensionality of latent spaces.

\subsection{Property Optimization}

Our objective is to generate a crystal that optimizes a specific property $\cP$ of the crystal structure, which is determined by the BB function $f^{\text{BB}}$.  The crystal exists within the input space $\cX$, leading us to define our optimization challenge as finding $\bm{x^*} \in \cX$ that maximizes $f^{\text{BB}}(\bm{x})$, expressed as:
\begin{equation}
\bm{x^*} = \arg\max_{\bm{x} \in \cX} f^{\text{BB}}(\bm{x}).
\end{equation}
However, due to the high dimensionality of $\cX$ and costly evaluation of $f^{\text{BB}}$, direct optimization in $\cX$ is impractical. Therefore, we instead perform optimization in the latent space $\cZ$, utilizing BO to navigate this space efficiently. The optimization problem in the latent space is therefore formulated as:
\begin{equation}
    \bm{z^*} = \arg\max_{\bm{z} \in \cZ} g(\bm{z}),
\end{equation}
where $g(\bm{z}) = f^{\text{BB}}(f^{\text{dec}}(\bm{z}))$ is the composition of the objective function with the decoder, mapping a latent space point back to the input space to evaluate its property $\mathcal{P}$ via the BB function.

\begin{figure}[!t]
  \centering
  \begin{minipage}{0.35\textwidth}
    \centering
    \includegraphics[width=\textwidth]{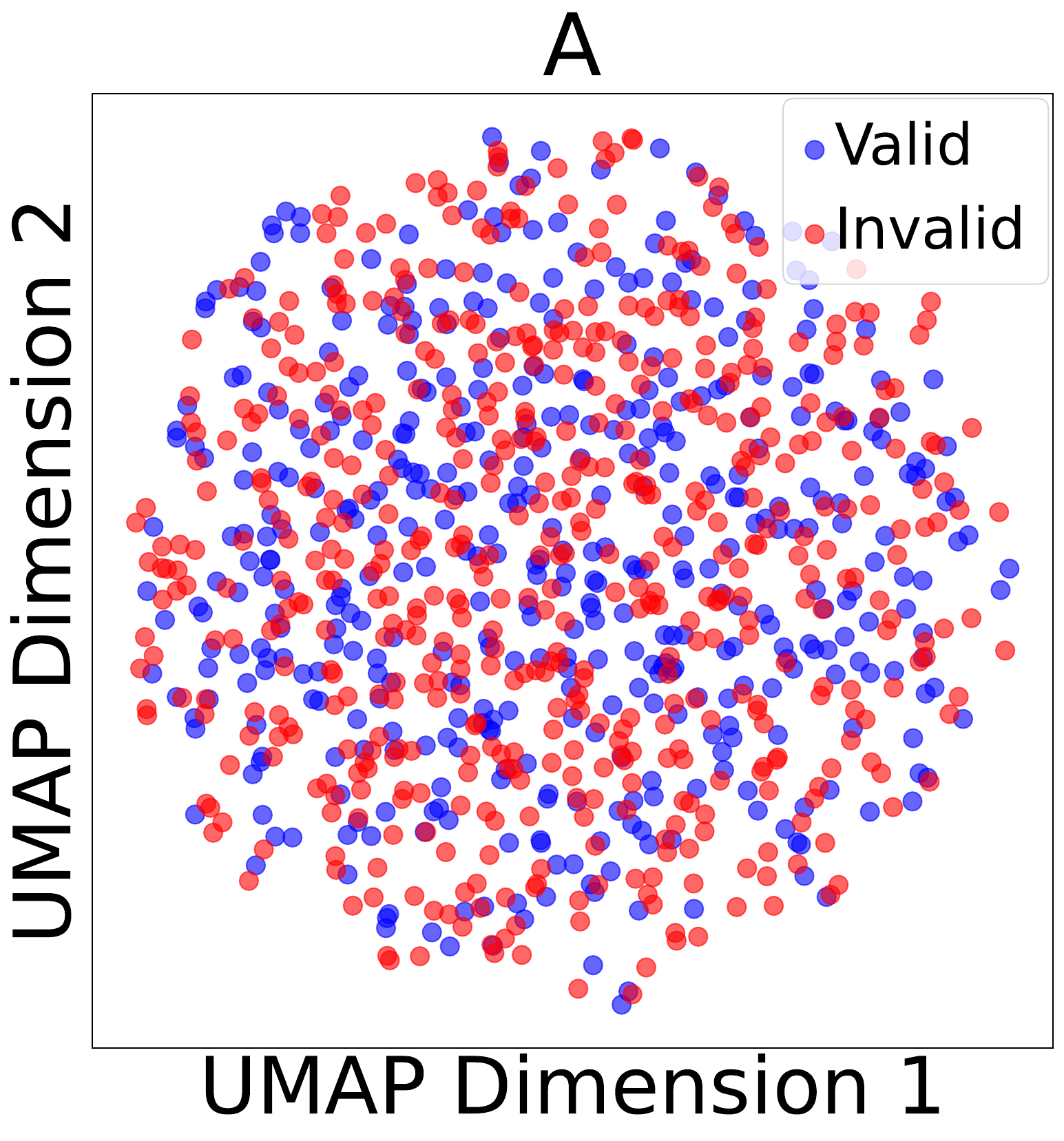}
    \label{fig:fig2a}
  \end{minipage}%
  \hspace{1.5pt}
  \begin{minipage}{0.35\textwidth}
    \centering
    \includegraphics[width=\textwidth]{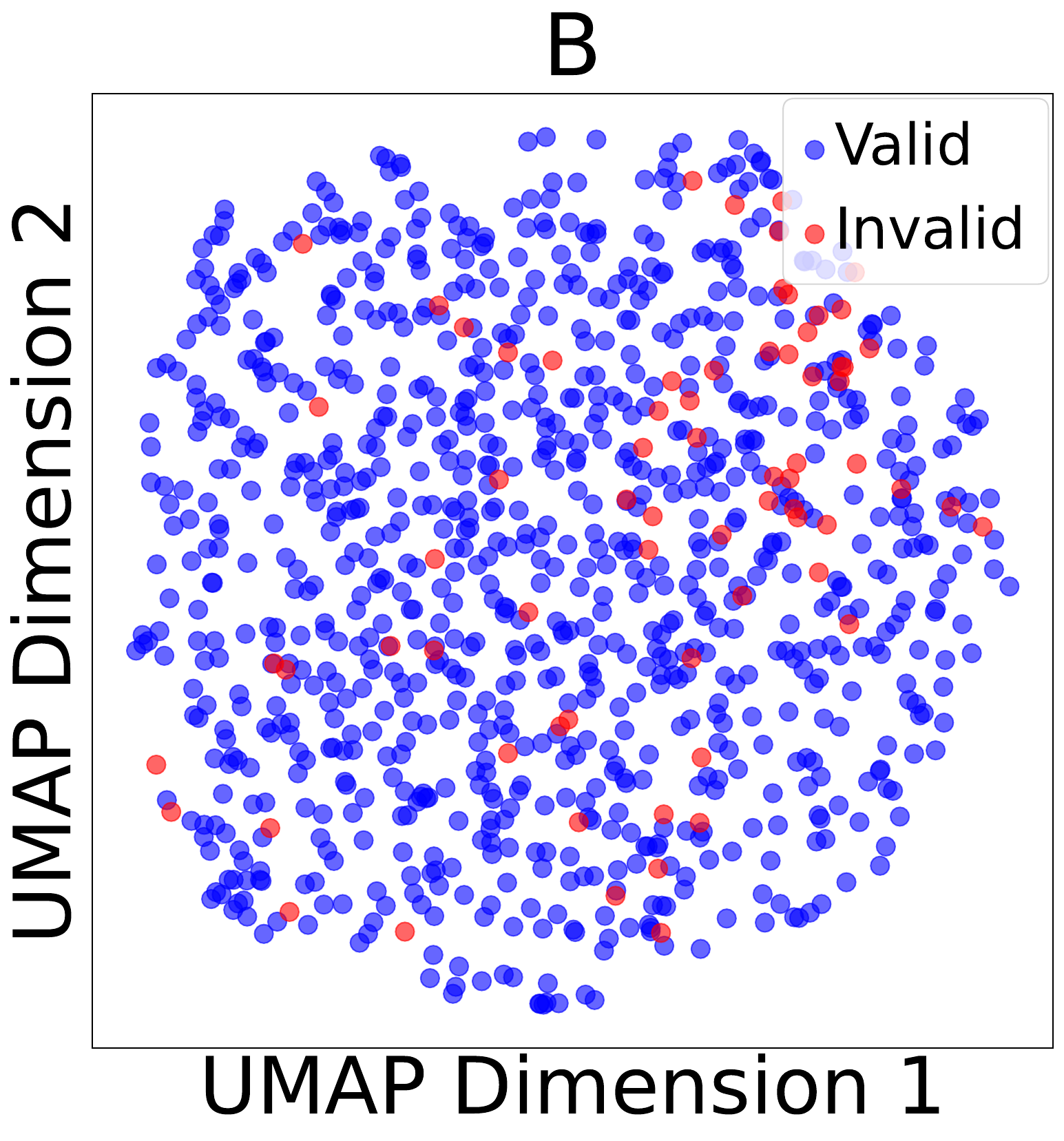}
    \label{fig:fig2b}
  \end{minipage}
    \caption{UMAP plots show the latent spaces for \textcolor{red}{invalid} and \textcolor{blue}{valid} generation. \textbf{A} details the FTCP-VAE model's latent space, mostly leading to invalid generations with only a 49\% validity rate. 
    \textbf{B} displays the Combined-VAE's material latent space within the Crystal-LSBO framework, highlighting its ability to generate valid crystals even from sparse regions, with a high validity rate of 93\%.
    } 
\label{fig:invalid_valid}
\end{figure}

\subsection{Latent Space Explorability}\label{sec:pd_ls_explorability}

Extensive exploration of the latent space and the ability to generate \emph{valid} crystal structures throughout this space is crucial for successful de novo crystal design. In the machine learning for materials science literature, the term ``validity'' is defined in various ways. In this study, a crystal structure is considered valid if it can be converted into a Crystallographic Information File (CIF), which is the standard format for crystal structures in materials science \cite{zhao2023physics}. The generated structure fails to be converted into a CIF if the lattice does not form a valid 3-dimensional structure, the coordinates are not within the correct range, or the lattice does not contain an element. 

LSBO's success hinges on its ability to explore the latent space, relying on the decoder's capability to accurately generate valid crystal structures from any latent point $\bm{z} \in \mathcal{Z}$. Therefore, the ability to generate valid materials from a broad region in the latent space is critical. Despite the wide range of use cases of VAEs in material generation tasks, existing models predominantly adhere to local search strategies, largely due to an underdeveloped capacity for wide-ranging exploration within their latent spaces, or due to using VAEs in combination with other non-generative models like AEs, U-Nets \cite{nohinverse, hoffmann2019data, court20203}. Among these studies, to our knowledge, FTCP-VAE stands out as the sole VAE-based method that does not integrate the VAE with non-VAE architectures, aligning it with the focus of our research. Hence, we are particularly interested in explorability of its latent space and generative capabilities. In order to make an evaluation, we trained an FTCP-VAE model\footnote{The code provided by the FTCP-VAE authors is used.} and performed 1000 generations using latent vectors drawn from the standard normal distribution, $\mathcal{N}(\bm{0}, \bm{I})$. Figure \ref{fig:invalid_valid}(A) shows a 2-dimensional UMAP plot of the model's distribution of valid and invalid generations in the latent space, highlighting challenges in consistently producing valid crystal structures, with only 49\% of the generated crystals meeting validity requirements. The FTCP-VAE, while utilizing a standalone VAE model to process crystal structure data, evidently struggles with the inherent complexity of such data, leading to limited generative and de novo design performance. On the other hand, Fig. \ref{fig:invalid_valid}(B) shows the latent space of the proposed Combined-VAE model (see \S\ref{sec:experiments} for more details), in which there are a much smaller number of invalid generations than the FTCP-VAE model in A.

\subsection{Latent Consistent-Aware LSBO}\label{sec:lca-lsbo}

In order to enhance exploration in the latent space of LSBO, we adopt the Latent Consistent-Aware LSBO (LCA-LSBO) \cite{boyar2024latent}, recently proposed in the field of organic molecule generation. Retraining of the VAE discussed above involves updating the VAE's training dataset with new instances from LSBO queries, and periodically retraining the VAE with this updated dataset. However, due to the expensive nature of BB function evaluations, it is unrealistic to expect a high number of new instances that are enough to bring a meaningful update to the VAE. LCA-LSBO uses label-free data augmentations in the latent space to address this challenge. Synthetic latent variables $\hat{\bm{z}}$ are generated from a probability distribution. These augmentations focus on regions of interest in the latent space that decode into instances with high property values or areas identified as promising by the AF of LSBO. Augmented variables are decoded, re-encoded, and discrepancies are penalized during VAE retraining. Namely, given augmented latent variable $\bm{\hat{z}}$, LCA-LSBO adds the additional term $||\bm{\hat{z}} - f^{\text{enc}}_\phi(f^{\text{dec}}_{\theta}(\bm{\hat{z}}))||^2$ to the objective function of VAE during retraining. This can be considered as the reconstruction of augmented latent variables at targeted regions. It provides the effective incorporation of data augmentations into the retraining process to address the problem of limited new data to use in retraining. Centered on the region of interest, these augmentations enable LSBO to more effectively explore and exploit promising regions in the latent space. The label-free nature of these augmentations and their targeted selection accelerate de novo discovery and enhance sample efficiency. For more detail on LCA-LSBO, see the Appendix and \cite{boyar2024latent}.

\section{Proposed Method}\label{sec:proposed_method}

In this section, we present the proposed Crystal-LSBO framework. We detail the VAE components of the Crystal-LSBO framework in \S\ref{sec:multi-vae}, and we detail our overall LSBO approaches in \S\ref{sec:proposed_lsbo}. Lastly, we detail the structure relaxation process for the generated crystals.

\begin{figure}[!t]
  \centering
  \begin{minipage}{0.6\textwidth}
    \centering
    \includegraphics[width=\textwidth]{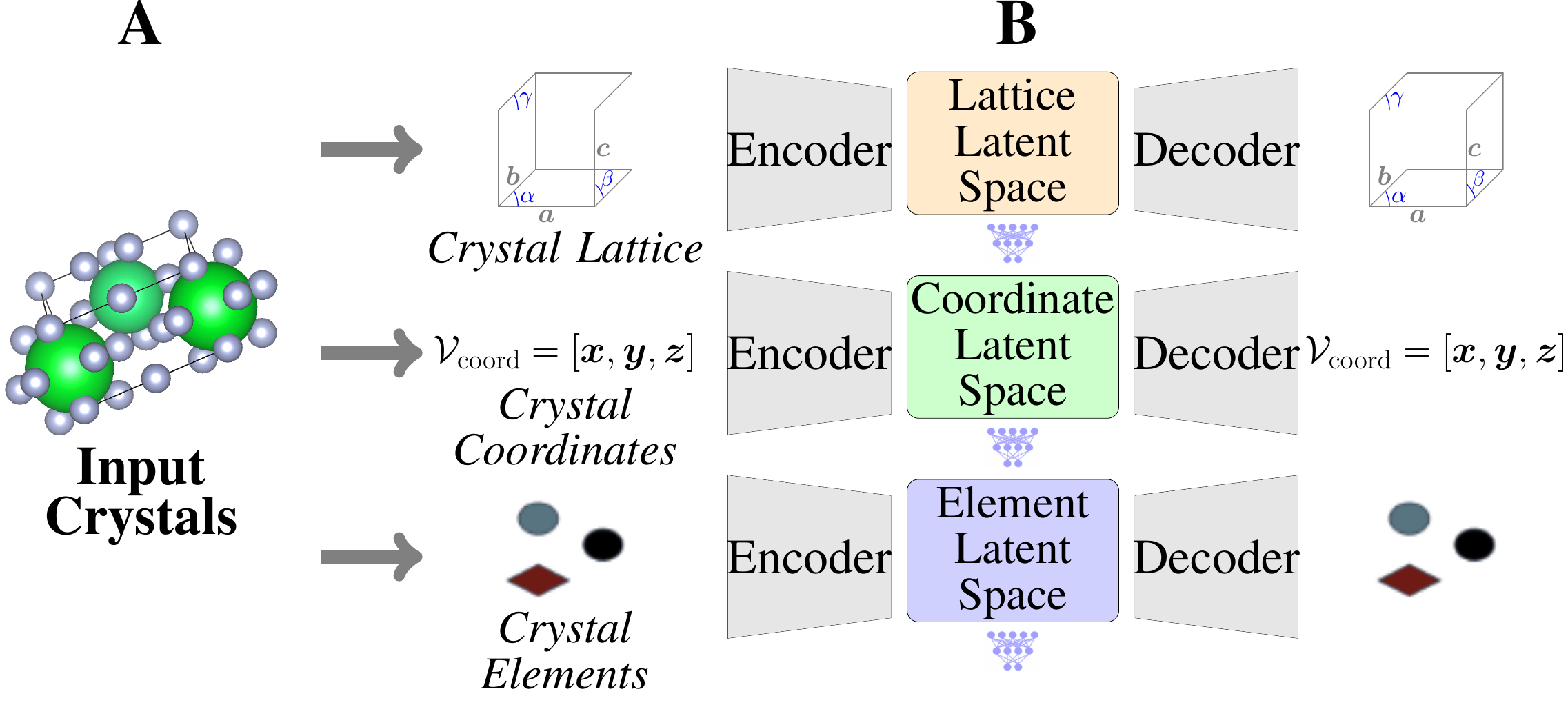}
    \label{fig:fig2a}
  \end{minipage}
  \begin{minipage}{0.48\textwidth}
    \centering
    \includegraphics[width=\textwidth]{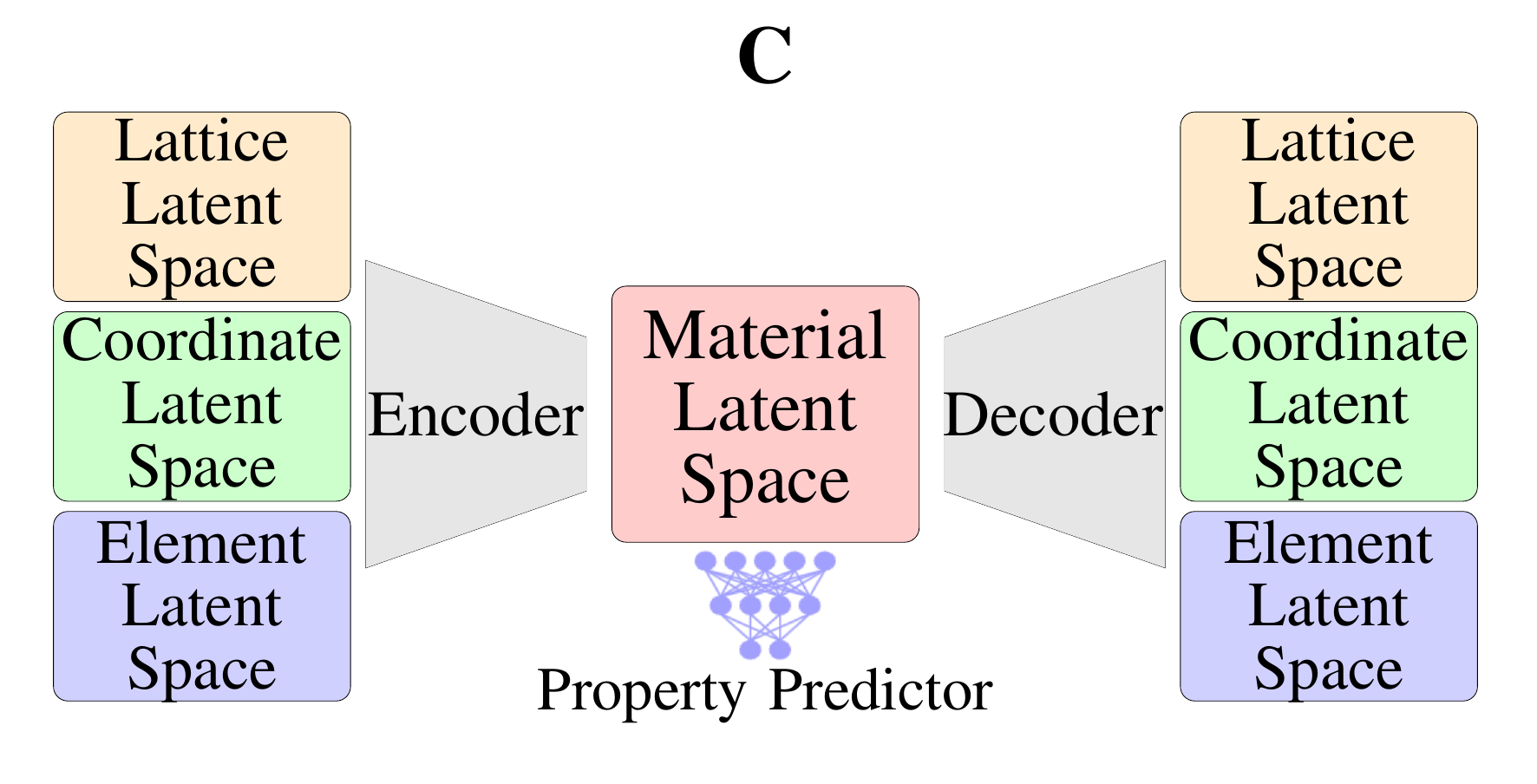}
    \label{fig:fig2b}
  \end{minipage}
 \caption{The VAE model architecture in the Crystal-LSBO framework operates as follows: Step \textbf{A} categorizes input crystals into Lattice, Coordinate, and Element parts. Step \textbf{B} trains separate VAEs for these parts, obtaining their latent representations. Step \textbf{C} merges these representations into a unified latent space through the Combined-VAE.}
 \label{fig:multi_vae}
\end{figure}

\subsection{VAE Components of the Crystal-LSBO Framework}\label{sec:multi-vae}
To address the problem of latent space explorability discussed in \S\ref{sec:pd_ls_explorability}, we adopted a multi-model approach, where each specialized VAE model focuses on a distinct aspect of the crystal structure to simplify the learning process. We used space group 1 representations of crystals for their simplicity and comprehensiveness, as all crystals can be represented in this form\footnote{Details regarding the selection of space group 1 and its implications are provided in \S \ref{sec:spacegroup1}.}. With such representation, we dissect the crystal information obtained from the CIF formats of crystals into three key components: lattice parameters (angles $[\alpha, \beta, \gamma]$ and cell lengths $[a, b, c]$), coordinates of the elements in crystals, and a one-hot encoding matrix for the elements alongside their occupancy information. These VAEs, referred to as Lattice-VAE parameterized by $f^{\text{enc}}_{\text{Latt-VAE}_{\phi}}$ and $f^{\text{dec}}_{\text{Latt-VAE}_{\theta}}$, Coordinate-VAE parameterized by $f^{\text{enc}}_{\text{Coord-VAE}_{\phi}}$ and $f^{\text{dec}}_{\text{Coord-VAE}_{\theta}}$, and Element-VAE parameterized by $f^{\text{enc}}_{\text{Elem-VAE}_{\phi}}$ and $f^{\text{dec}}_{\text{Elem-VAE}_{\theta}}$, are trained on these components. Subsequently, the latent representations obtained from these VAEs are used to train another model, referred to as Combined-VAE, parameterized by $f^{\text{enc}}_{\text{Combined-VAE}_{\phi}}$ and $f^{\text{dec}}_{\text{Combined-VAE}_{\theta}}$. This model synthesizes the diverse input components into a cohesive material latent space. Each VAE is co-trained with a neural network-based property predictor (PP) model to organize the distribution of the crystals in the latent space by their property values $\bm{y}$. The model architecture is provided in Fig. \ref{fig:multi_vae}. The objective function for each VAE in the Crytal-LSBO framework is defined as:
\begin{equation}
\mathcal{L}^{\text{VAE}}_{\text{Crystal-LSBO}}(\theta, \phi; \bm{x}) = \mathbb{E}_{q_{\phi}(\bm{z} | \bm{x})} [\log p_{\theta}(\bm{x} | \bm{z})] - \beta \, D_{\text{KL}}(q_{\phi}(\bm{z} | \bm{x}) \| p(\bm{z})) - w ||\bm{y}- \text{PP}_{\psi}(\bm{y}|\bm{z}))||^2,
\end{equation}
where $w$ is the weight assigned to the property predictor.\footnote{Information on the model architectures of the VAEs, the PP, and training details are provided in the Appendix.}

\begin{figure}
 \centering
 \includegraphics[width=0.75\textwidth]{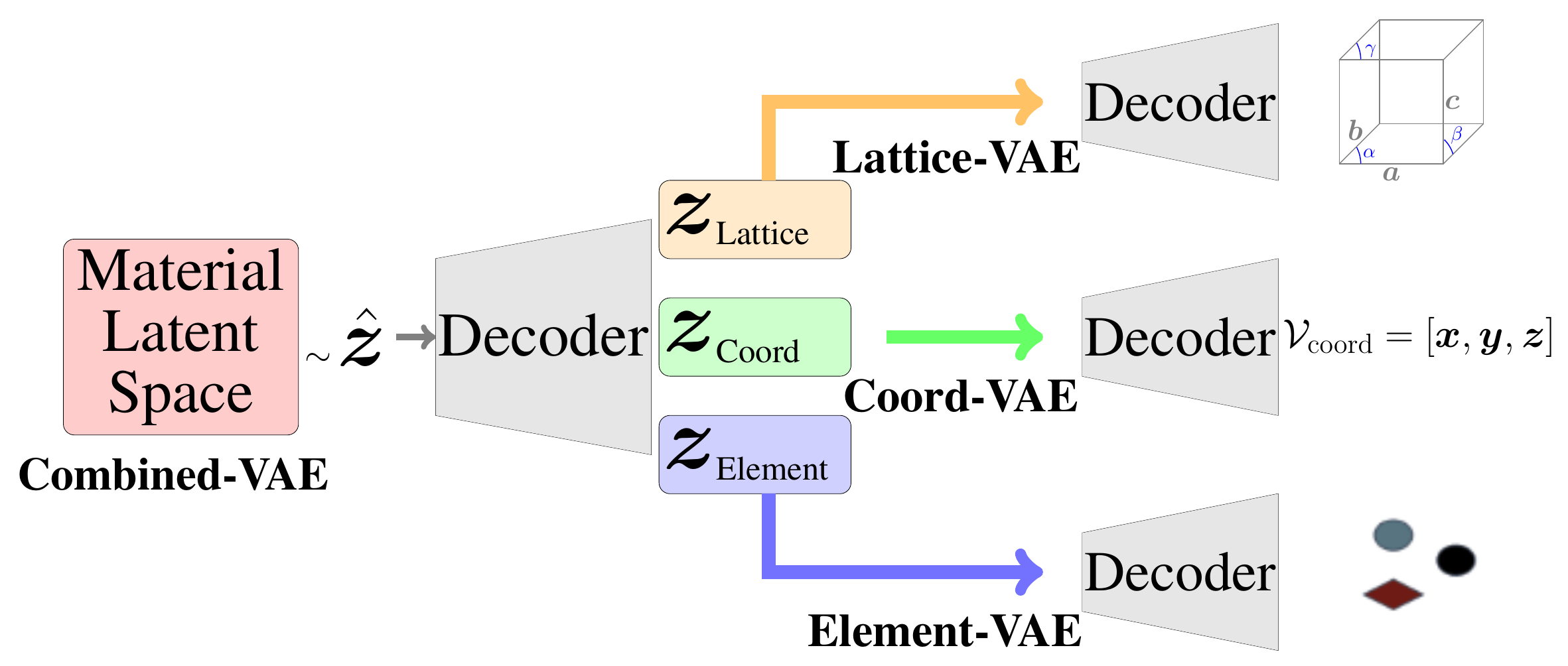}
 \caption{In the Crystal-LSBO framework, crystal generation unfolds as follows: First, latent variables $\hat{\bm{z}}$ are sampled from the material latent space. The Combined-VAE's decoder then produces specific latent representations for each VAE. Next, these representations are used by the Lattice, Coordinate, and Element-VAE decoders to generate the respective crystal components. The final structure is assembled from these components. During LSBO, this generation process is guided by the AF of BO.
 }
 \label{fig:multi_vae_gen}
\end{figure}

\subsection{Optimization with LSBO in the Crystal-LSBO Framework}\label{sec:proposed_lsbo}

In this subsection, we describe the LSBO algorithms considered in the Crystal-LSBO framework.

\subsubsection{Crystal-Standard-LSBO}

We start by detailing how the standard LSBO algorithm, described in \S\ref{sec:2.2}, is incorporated into our Crystal-LSBO framework, which we refer to as Crystal-Standard-LSBO. In this setup, the latent space created by the Combined-VAE establishes the search domain for Crystal-Standard-LSBO. Within this domain, a GP model navigates the landscape, with the sampling process guided by the AF. When the AF algorithm is maximized at the region $\bm{z^*}$, the Combined-VAE's decoder transforms it into an output matrix that holds the latent variables for the Lattice, Coordinate, and Element parts. These latent variables $\bm{z}_{\text{Lattice}}^*,\bm{z}_{\text{Coordinate}}^*, \bm{z}_{\text{Element}}^*$ are then input into their respective VAEs to generate the corresponding components of the crystal structure, as demonstrated in Fig. \ref{fig:multi_vae_gen}. The subsequent step involves assembling these components to form a complete crystal structure, which is then assessed using the BB function. Following this evaluation, the GP model is updated with the new BB function value and $\bm{z^*}$, and AF uses the updated GP to find the next query point. The Crystal-Standard-LSBO approach is outlined in Algorithm 1.

\begin{algorithm}
    \caption{Crystal-Standard-LSBO}
    \begin{algorithmic}[1]
      \Require Trained Crystal-LSBO VAEs, Labeled instances $\mathcal{L}$, BB Function $f^{\text{BB}}$, Experiment count $J$
      \For{$t = 1$ to $J$}
        \State Fit GP model using $\{(f_\phi^{\text{enc}}(x_i), y_i)\}_{i \in \mathcal{L}}$
        \State Find $\bm{z}^* \leftarrow \arg\max_{\bm{z} \in \mathcal{Z}} \hat{f}^{\text{AF}}(\bm{z}_i)$
        \State Separate $f^{\text{dec}}_{\text{Combined-VAE}_{\theta}}(\bm{z}^*)$ into $\bm{z}_{\text{Lattice}}^*, \bm{z}_{\text{Coordinate}}^*, \bm{z}_{\text{Element}}^*$
        \State  $\bm{x}_{\text{Lattice}}^* \gets f^{\text{dec}}_{\text{Latt-VAE}_{\theta}}(\bm{z}_{\text{Lattice}}^*)$
        \State $\bm{x}_{\text{Coordinate}}^* \gets f^{\text{dec}}_{\text{Coord-VAE}_{\theta}}(\bm{z}_{\text{Coordinate}}^*)$
        \State $\bm{x}_{\text{Element}}^* \gets f^{\text{dec}}_{\text{Elem-VAE}_{\theta}}(\bm{z}_{\text{Element}}^*)$
        \State Form the crystal by combining $\hat{\bm{x}}^* = [\bm{x}_{\text{Lattice}}^* \cup \bm{x}_{\text{Coordinate}}^* \cup \bm{x}_{\text{Element}}^*]$
         \State Evaluate label: $y^* = f^{\text{BB}}(\hat{\bm{x}}^*)$
        \State Update $\mathcal{L} \leftarrow \mathcal{L} \cup \{(\hat{x}^*, y^*)\}$ 
      \EndFor
    \end{algorithmic}
\end{algorithm}

\subsubsection{Crystal-LCA-LSBO}\label{sec:lca-lsbo-ex}

As discussed in \S\ref{sec:lca-lsbo}, LCA-LSBO employs label-free data augmentations within the latent space to facilitate rapid updates to the model after each retraining. In this section, we explain the implementation of LCA-LSBO within our Crystal-LSBO framework, referred to as Crystal-LCA-LSBO. In Crystal-LCA-LSBO, different than Crystal-Standard-LSBO, each of the four distinct VAE models undergoes periodic retraining. This retraining is guided by the identification of regions of interest based on the target property values of the crystals. Specifically, when the target property values of queried instances exceed a defined threshold \( \tau \), such regions are deemed as the regions of interest, and data augmentations in the latent space are generated to explore these promising areas. Specifically, when a query point results in the generation of a crystal whose target property value exceeds a predefined threshold, the latent variables used to generate this crystal—$\bm{z}_{\text{Lattice}}^*$ in Lattice-VAE, $\bm{z}_{\text{Coordinate}}^*$ in Coordinate-VAE, and $\bm{z}_{\text{Element}}^*$ in Element-VAE models—are designated as regions of interest in latent space of their respective models. 
The next steps include simultaneous retraining of the Lattice-VAE, Coordinate-VAE, and Element-VAE. This process involves generating synthetic latent variables in the neighborhood of the regions of interest, and retraining with the penalization term for augmented latent variable reconstructions, as detailed in \S \ref{sec:lca-lsbo} and \S \ref{sec:app1}. Specifically, we consider a normal distribution $p^{\rm ref} \sim \mathcal{N}(\bm{\mu}^{\rm ref}, \sigma^{\rm ref})$, which is centered around latent variables that exceed a specific property value threshold, and use randomly generated latent variables from this distribution as augmented latent variables (The distributions in the latent spaces of Lattice, Coordinate, and Element VAE are respectively referred to as $p^{\rm ref}_{\rm Lattice}$, $p^{\rm ref}_{\rm Coordinate}$, and $p^{\rm ref}_{\rm Element}$).
Next, the Combined-VAE is also retrained using the latent representations of the training instances obtained from the retrained individual VAEs. After Combined-VAE is retrained, the process continues with finding the next query point via AF and its evaluation. The algorithm of Crystal-LCA-LSBO is provided in Algorithm 2.

\begin{algorithm}
    \caption{Crystal-LCA-LSBO}
    \begin{algorithmic}[1]
        \Require Trained Crystal-LSBO VAEs, Unlabeled instances: $\mathcal{U}$, Labeled instances: $\mathcal{L}$, $f^{\text{AF}}(\bm{z})$, BB Function $f^{\text{BB}}$, Experiment count $J$, $\sigma^{\text{ref}}$,  Sample size of $\hat{\bm{z}}$: $N^*$, Region of Interest threshold $\tau$
        \For{$j = 1, \dots, J$}
            \State Fit GP model using $\{(f_\phi^{\text{enc}}(x_i), y_i)\}_{i \in \mathcal{L}}$
            \State Find $\bm{z}^* \leftarrow \arg\max_{\bm{z} \in \mathcal{Z}} \hat{f}^{\text{AF}}(\bm{z})$
            \State Separate $f^{\text{dec}}_{\text{Combined-VAE}_{\theta}}(\bm{z}^*)$ into $\bm{z}_{\text{Lattice}}^*, \bm{z}_{\text{Coordinate}}^*, \bm{z}_{\text{Element}}^*$.
        \State  $\bm{x}_{\text{Lattice}}^* \gets f^{\text{dec}}_{\text{Latt-VAE}_{\theta}}(\bm{z}_{\text{Lattice}}^*)$
        \State $\bm{x}_{\text{Coordinate}}^* \gets f^{\text{dec}}_{\text{Coord-VAE}_{\theta}}(\bm{z}_{\text{Coordinate}}^*)$
        \State $\bm{x}_{\text{Element}}^* \gets f^{\text{dec}}_{\text{Elem-VAE}_{\theta}}(\bm{z}_{\text{Element}}^*)$
            \State Form the crystal by combining $\hat{\bm{x}}^* = [\bm{x}_{\text{Lattice}}^* \cup \bm{x}_{\text{Coordinate}}^* \cup \bm{x}_{\text{Element}}^*]$
            \State Evaluate label: $y^* = f^{\text{BB}}(\hat{\bm{x}}^*)$
            \State Update $\mathcal{L} \leftarrow \mathcal{L} \cup \{(\hat{\bm{x}}^*, y^*)\}$ 
             \If{$|y^*| > |\tau| $}
                 \State Set $\bm{\mu}^{\text{ref}}_{\text{Lattice}} \leftarrow \bm{z}_{\text{Lattice}}^*$ and obtain $\{\hat{\bm{z}}^i_{\text{Lattice}}\}_{i=1}^{N^*} \sim p^{\text{ref}}_{\text{Lattice}}$
                 \State Set $\bm{\mu}^{\text{ref}}_{\text{Coordinate}} \leftarrow \bm{z}_{\text{Coordinate}}^*$ and obtain $\{\hat{\bm{z}}^i_{\text{Coordinate}}\}_{i=1}^{N^*} \sim p^{\text{ref}}_{\text{Coordinate}}$
                 \State Set $\bm{\mu}^{\text{ref}}_{\text{Element}} \leftarrow \bm{z}_{\text{Element}}^*$ and obtain $\{\hat{\bm{z}}^i_{\text{Element}}\}_{i=1}^{N^*} \sim p^{\text{ref}}_{\text{Element}}$
                 \State ReTrain Lattice-VAE with $\{(\mathcal{L} \cup \mathcal{U})_{\text{Lattice}}\}$ and $\{\hat{\bm{z}}^i_{\text{Lattice}}\}$
                 \State ReTrain Coordinate-VAE with $\{(\mathcal{L} \cup \mathcal{U})_{\text{Coordinate}}\}$ and $\{\hat{\bm{z}}^i_{\text{Coordinate}}\}$
                 \State ReTrain Element-VAE with $\{(\mathcal{L} \cup \mathcal{U})_{\text{Element}}\}$ and $\{\hat{\bm{z}}^i_{\text{Element}}\}$
                 \State $\bm{z}_{\text{Lattice}} \gets f^{\text{enc}}_{\text{Latt-VAE}_{\phi}}(\{(\mathcal{L} \cup \mathcal{U})\}_{\text{Lattice}})$ 
                 \State $\bm{z}_{\text{Coordinate}} \gets f^{\text{enc}}_{\text{Coord-VAE}_{\phi}}(\{(\mathcal{L} \cup \mathcal{U})\}_{\text{Coordinate}})$ 
                 \State $\bm{z}_{\text{Element}} \gets f^{\text{enc}}_{\text{Elem-VAE}_{\phi}}(\{(\mathcal{L} \cup \mathcal{U})\}_{\text{Element}})$ 
                 \State ReTrain Combined-VAE with $\bm{z}_{\text{Coordinate}}, \bm{z}_{\text{Coordinate}}, \bm{z}_{\text{Element}}$
            \EndIf
        \EndFor
    \end{algorithmic}
\end{algorithm}

\textbf{Post-Processing with M3GNet Structural Relaxation:} After completing the Crystal-LSBO algorithms, we use M3GNet \cite{Chen2022}, a popular machine learning model for structural relaxation. M3GNet refines crystals with desired properties identified by LSBO. However, success in structural relaxation is not guaranteed; issues like non-convergence or elemental overlaps can cause failures. Thus, the success rate of relaxation offers insights into the generative model's ability to produce stable crystals. 

\section{Experiments}\label{sec:experiments}

In our study, we used a dataset of ternary crystals with primitive cell representations sourced from the Materials Project \cite{Jain2013}, focusing on structures with no more than 40 sites, energy above hull of less than 0.08 eV/atom. Band gap and formation energy values of the crystals serve as target outputs for the property prediction models in each VAE. For LSBO experiments, we employed a GP with an RBF kernel featuring Automatic Relevance Determination to dynamically adjust the lengthscale for each latent dimension within the search space of $[-3, 3]$. Expected Improvement was utilized as the AF.

\subsection{Optimizing Electronegativity for Model Selection}\label{sec:lsbo-model-selection}

It is difficult to objectively determine the optimal latent dimensions and hyperparameters for LSBO due to inherent trade-offs: lower latent dimensions increase VAE reconstruction errors, while higher latent dimensions make BO more challenging. To address this, we utilize a surrogate optimization problem based on \emph{electronegativity}, a property that influences a crystal’s stability and reactivity. This property can be efficiently evaluated using the pymatgen library \cite{pymatgen}. We then conduct LSBO experiments using VAEs trained with different configurations. It is important to note that achieving optimal results with the surrogate problem does not necessarily guarantee similar outcomes with the actual target problem. This approach, however, assists in objectively identifying feasible latent dimensions and hyperparameters. For simplicity, we set the Lattice-VAE model to use a 3-dimensional latent space, while the Element-VAE and Coordinate-VAE models utilized 16-dimensional latent spaces with fixed $\beta$ and $w$ values.
The Combined-VAE, incorporating learned latent representations from these models, is trained across latent dimensions of $\{16, 20, 24\}$ and hyperparameters $\beta$ and $w$ ranging from $\{0.01, 0.1, 1, 5\}$, creating 48 model variations. GP is trained on 100 instances with the highest electronegativity values from our dataset, and LSBO experiments for each Combined-VAE model are conducted using the Crystal-Standard-LSBO method outlined in Algorithm 1, across 100 iterations and 10 different seeds.

\begin{figure}[!t]
  \centering
  \begin{minipage}{0.35\textwidth}
    \centering
    \includegraphics[width=\textwidth]{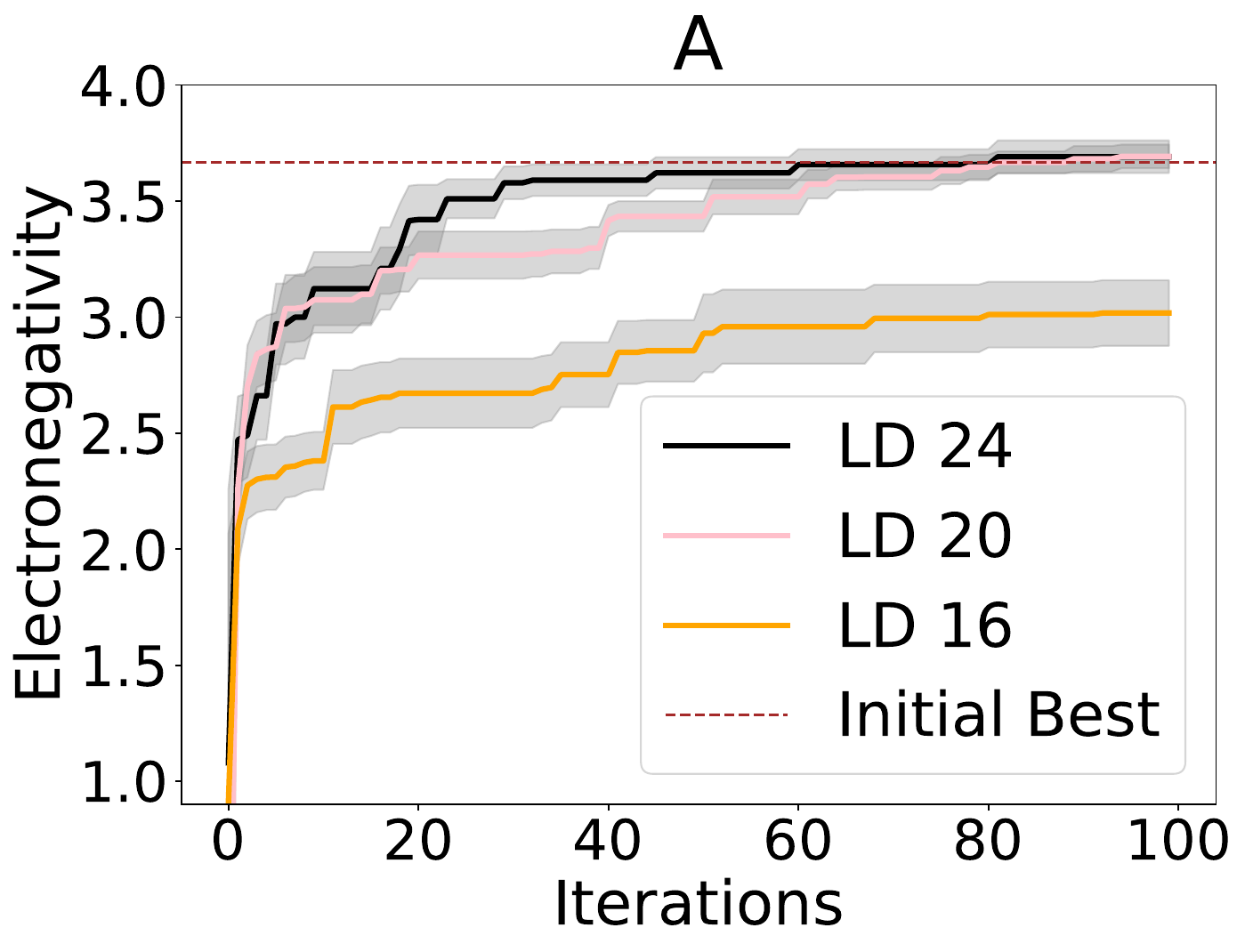}
    \label{fig:fig2a}
  \end{minipage}
  \hspace{0.3cm}
  \begin{minipage}{0.61\textwidth}
    \centering
    \includegraphics[width=\textwidth]{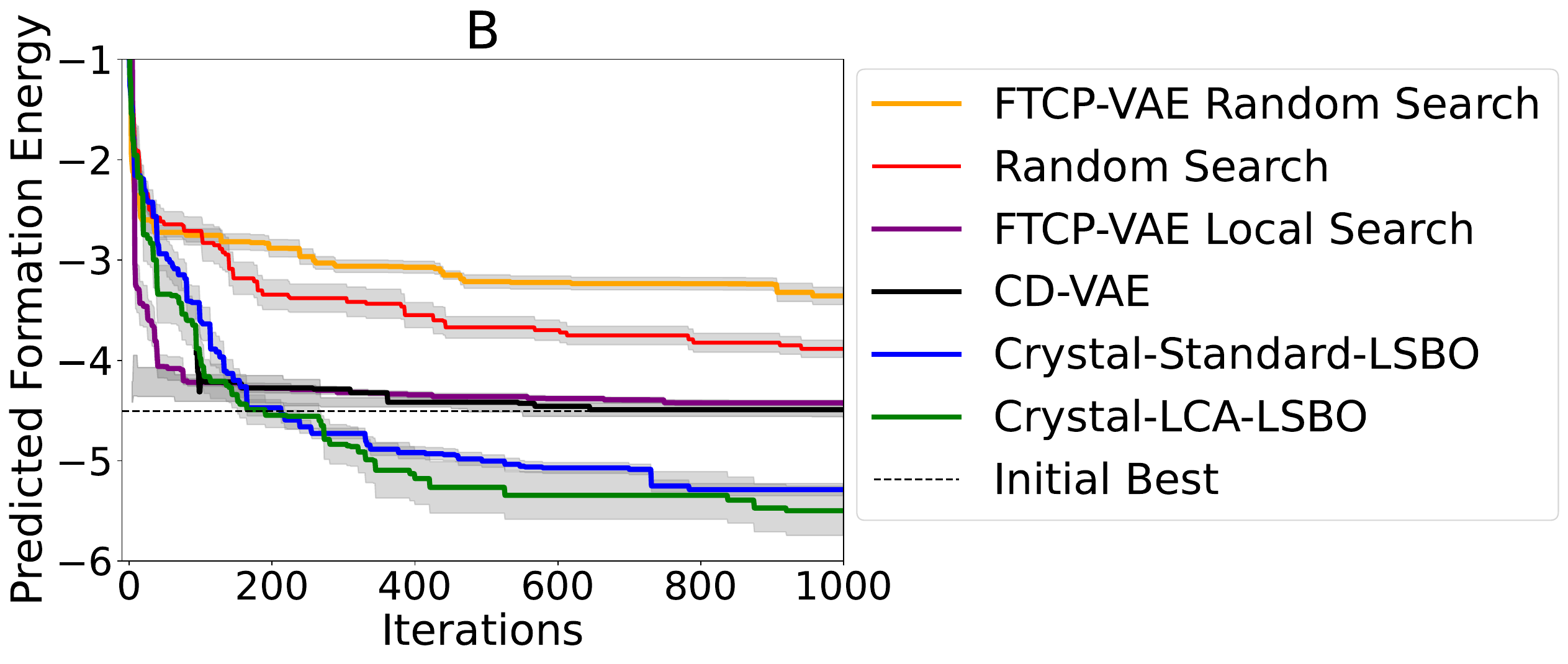}
    \label{fig:fig2b}
  \end{minipage}
    \caption{Panel \textbf{A} showcases the outcomes of using the Crystal-Standard-LSBO method for the LSBO task focused on designing de novo crystals with enhanced electronegativity values. The 20-dimensional latent space model emerged as optimal. Panel \textbf{B} focuses on the optimization of predicted formation energies, comparing the performance of Crystal-Standard-LSBO, Crystal-LCA-LSBO, CD-VAE, Random Search, FTCP-VAE Random Search, and FTCP-VAE Local Search. Crystal-LCA-LSBO significantly outperformed the others, while Crystal-Standard-LSBO also showed effective results. 
    } 
\label{fig:en_fe_lsbo_results}
\end{figure}

Our selection criteria favored models with the lowest latent dimension among the top-performing options. In Fig. \ref{fig:en_fe_lsbo_results}(A), we showcase the top-performing Combined-VAEs across 16, 20, and 24-dimensional latent spaces, in which the average of the results from different seeds and the standard errors are displayed (Higher is the better). We found that the best performing 16-dimensional model lagged in LSBO tasks, while the 20 and 24-dimensional models exhibited similar efficacy, in which both can generate crystals with higher electronegativity values than the highest value among known crystals in our dataset within 100 iterations. Following our selection criteria, we chose the Combined-VAE with 20-dimensional latent space, which is trained with $\beta = 0.01$ and $w = 5$.

\textbf{Validity of Generations:} In \S\ref{sec:pd_ls_explorability}, we discussed the problem of latent space explorability and pointed out that the existing models suffer from invalid generations and provided an example using the FTCP-VAE model, which is trained on the same dataset as ours. Using the selected model, we conducted a similar analysis and provided the results in Fig. \ref{fig:invalid_valid}(B). In this setting, 1000 latent vectors are sampled from $\mathcal{N}(\bm{0}, \bm{I})$, and crystals are generated as demonstrated in Fig. \ref{fig:multi_vae_gen}. Among 1000 generations, 930 of them were successfully converted into a CIF format, resulting in a 93\% validity rate. It indicates a clear improvement in the validity rate over FTCP-VAE, showcasing the model's capacity to generate valid instances from a wide range in the latent space.

\subsection{Optimizing Formation Energy}\label{sec:opt_fe}

This section focuses on our main goal: designing crystals with optimal formation energy values.  Due to the high resource demands of exact formation energy calculations, we employed a proxy model as our BB function for estimating the formation energies of generated crystals during LSBO. We utilized an Xgboost \cite{chen2016xgboost} model trained on our crystal structure data and their formation energies, achieving a test Mean Squared Error (MSE) of 0.04 in predicting formation energies. Using the selected model in \S\ref{sec:lsbo-model-selection}, we implemented the proposed Crystal-Standard-LSBO and Crystal-LCA-LSBO algorithms\footnote{Details on hyperparameter selection for Crystal-LCA-LSBO are provided in the Appendix.}. The GP for each of the LSBO experiments was trained on the top 100 instances with the lowest predicted formation energies. To benchmark the effectiveness of the LSBO methods, we implemented a random search strategy using both our VAEs and the FTCP-VAE model, maintaining the same search bounds as those used in the LSBO experiments. Additionally, we adopted the local search methodology proposed by the FTCP-VAE authors, which involves sampling from the vicinity of the latent representation of the crystal with the best-predicted formation energy. We also applied the CD-VAE methodology proposed by \cite{xie2021crystal}, which uses a crystal diffusion model and incorporates a VAE to condition the generation process. Their approach uses gradient-based optimization, guided by a property predictor to steer the generation towards desired properties\footnote{The code provided by the CD-VAE authors was used.}. Each method was evaluated over 1,000 iterations across 10 different seeds.

Figure \ref{fig:en_fe_lsbo_results}(B) presents the average results and standard errors across different seeds, with lower values indicating better performance. Both the Crystal-Standard-LSBO and Crystal-LCA-LSBO methods consistently outperformed initial benchmarks by identifying crystals with lower predicted formation energies across all seeds. In contrast, the FTCP-VAE Local Search approach only marginally surpassed the initial benchmarks in 2 of the 10 seeds, failing to deliver consistent improvements. Furthermore, while random search strategies using our model and the FTCP-VAE model failed to provide improvements, our model demonstrated comparatively better results. CD-VAE results, on the other hand, show clear improvements over random search approaches and slightly outperform the FTCP-VAE Local Search, which focuses on sampling near the best-known values. However, CD-VAE struggles to consistently surpass the initial benchmark values and is outperformed by the Crystal-LSBO methodologies. In each optimization experiment, CD-VAE generates 10 candidates for each of the 1000 iterations and selects the best one using an additional predictor model. This results in CD-VAE sampling 10,000 materials per experiment, compared to 1,000 samples in other models. While this extensive sampling contributes to improved performance over FTCP-VAE, it also highlights the inherent complexity and inefficiency of the search and optimization process in CD-VAE. These findings underscore the efficiency of our Crystal-LSBO methodology in exploring the latent space to identify crystals that meet the desired criteria.



Among the two LSBO approaches, Crystal-LCA-LSBO proved particularly effective, consistently finding crystals with lower predicted formation energies. Table 1 consolidates the three lowest predicted formation energies for each method across all seeds, with the best results achieved by Crystal-LCA-LSBO, further highlighting its superior performance.

\begin{figure}[ht]
  \centering
  \begin{subfigure}{0.19\textwidth}
    \centering
    \includegraphics[width=\linewidth]{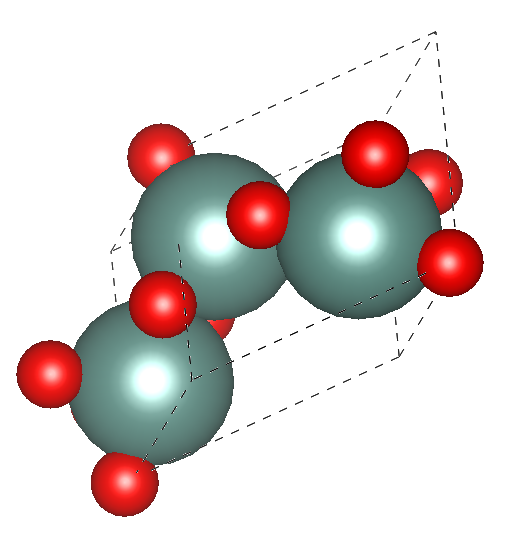}
    \caption{Y3 O4}
    \label{fig:sub2}
  \end{subfigure}
  \begin{subfigure}{0.19\textwidth}
    \centering
    \includegraphics[width=\linewidth]{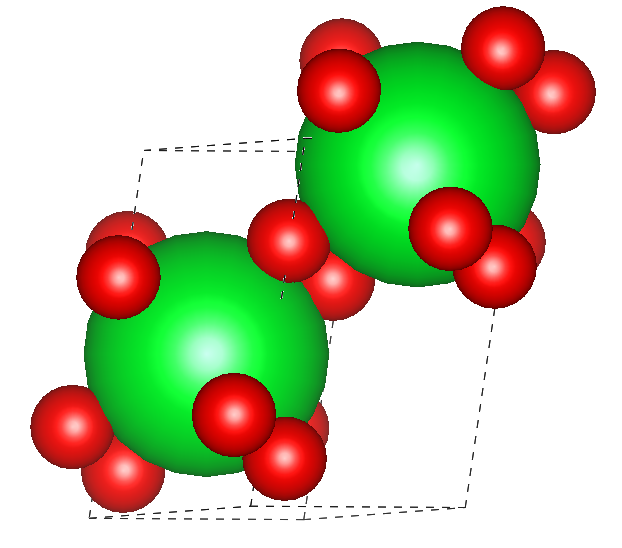}
    \caption{Sr2 O4}
    \label{fig:sub2}
  \end{subfigure}
  \begin{subfigure}{0.21\textwidth}
    \centering
    \includegraphics[width=\linewidth]{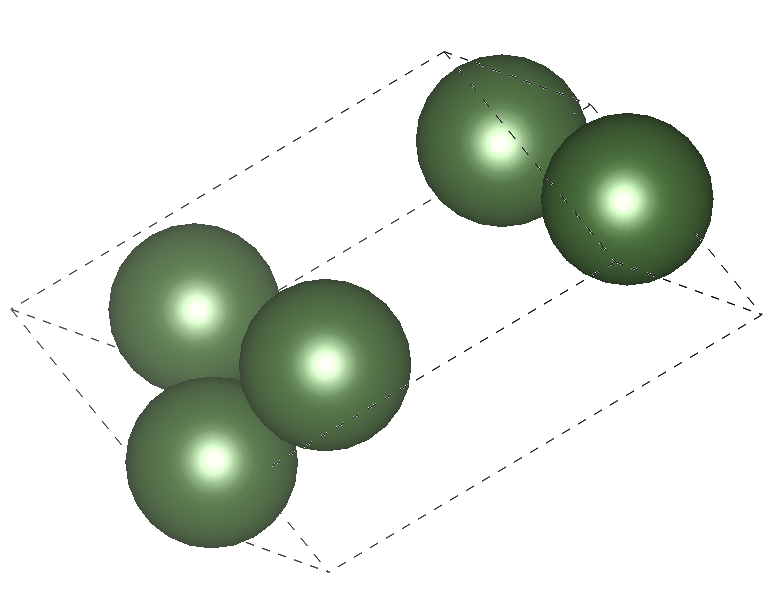}
    \caption{Er5}
    \label{fig:sub3}
  \end{subfigure}
  \begin{subfigure}{0.17\textwidth}
    \centering
    \includegraphics[width=\linewidth]{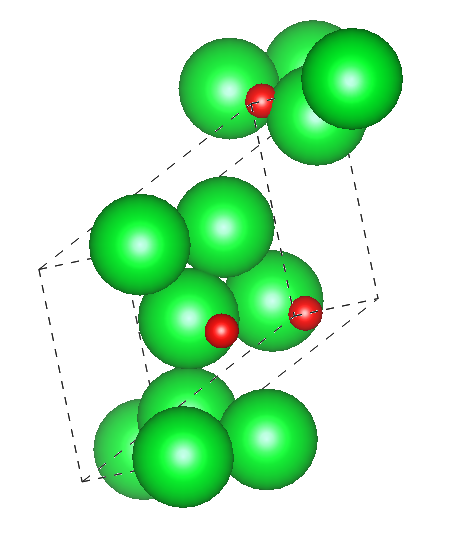}
    \caption{Sr4 O2}
    \label{fig:sub4}
  \end{subfigure}
  \caption{Examples of de novo crystals and their chemical compositions generated by Crystal-LCA-LSBO, showcasing diverse configurations.} 
  \label{fig:m3gnet_crystals}
\end{figure}

As detailed in \S\ref{sec:proposed_method}, we apply structural relaxation to the structures generated through LSBO. From our Crystal-LCA-LSBO experiments, we identified 38 crystals with predicted formation energies lower than the initial benchmarks. These crystal structures are then processed using the M3GNet model for structural relaxation. Among these, 36 successfully underwent the structural relaxation process, demonstrating a success rate of 94.7\%. Figure \ref{fig:m3gnet_crystals} illustrates examples from these crystals.

\begin{table}[ht]
\centering
\caption{Comparison of TOP 3 crystals with lowest predicted formation energies.}
\begin{tabular}{c|c|c|c}
\hline
\textbf{Method}  & $\mathbf{1^{st}}$ & $\mathbf{2^{nd}}$ & $\mathbf{3^{rd}}$  \\ \hline
Database & -4.50 & -4.43 & -4.14 \\
FTCP-VAE Random Search & -3.93 & -3.59 & -3.59 \\
Random Search & -4.26 & -4.25 & -4.17 \\
FTCP-VAE Local Search & -4.57 & -4.56 & -4.46 \\
CD-VAE & -4.62 & -4.57 & -4.48 \\
Crystal-Standard-LSBO & -6.05 & -5.53 & -5.45 \\
\textbf{Crystal-LCA-LSBO} & \textbf{-6.87} & \textbf{-6.49} & \textbf{-6.31} \\
\end{tabular}
\end{table}

\section{Conclusion}\label{sec:conclusion}

In this study, we demonstrated the effectiveness of the Crystal-LSBO framework for generative modeling and the de novo design of crystal structures. Our approach produced explorable latent spaces, enabling the pioneering application of LSBO algorithms to crystal design. The Crystal-Standard-LSBO method, enhanced by data augmentations in the latent space through the Crystal-LCA-LSBO method, effectively explored and exploited these spaces. This led to the successful discovery of crystals with target properties, showcasing the potential of Crystal-LSBO to significantly enhance automated crystal design and impact material discovery across various industries.

\bibliography{references}

\begin{thebibliography}{26}
\providecommand{\natexlab}[1]{#1}
\providecommand{\url}[1]{\texttt{#1}}
\expandafter\ifx\csname urlstyle\endcsname\relax
  \providecommand{\doi}[1]{doi: #1}\else
  \providecommand{\doi}{doi: \begingroup \urlstyle{rm}\Url}\fi

\bibitem[Boyar and Takeuchi(2024)]{boyar2024latent}
Onur Boyar and Ichiro Takeuchi.
\newblock {Latent Space Bayesian Optimization With Latent Data Augmentation for Enhanced Exploration}.
\newblock \emph{Neural Computation}, pages 1--33, 09 2024.
\newblock ISSN 0899-7667.
\newblock \doi{10.1162/neco_a_01708}.
\newblock URL \url{https://doi.org/10.1162/neco\_a\_01708}.

\bibitem[Brock et~al.(2016)Brock, Hahn, Wondratschek, M{\"u}ller, Shmueli, Prince, Authier, Kopsk{\`y}, Litvin, Arnold, et~al.]{brock2016international}
Carolyn~Pratt Brock, T~Hahn, H~Wondratschek, U~M{\"u}ller, U~Shmueli, E~Prince, A~Authier, V~Kopsk{\`y}, D~Litvin, E~Arnold, et~al.
\newblock International tables for crystallography volume a: Space-group symmetry, 2016.

\bibitem[Chen and Ong(2022)]{Chen2022}
Chi Chen and Shyue~Ping Ong.
\newblock A universal graph deep learning interatomic potential for the periodic table.
\newblock \emph{Nature Computational Science}, 2\penalty0 (11):\penalty0 718--728, 11 2022.
\newblock ISSN 2662-8457.
\newblock \doi{10.1038/s43588-022-00349-3}.
\newblock URL \url{https://doi.org/10.1038/s43588-022-00349-3}.

\bibitem[Chen and Guestrin(2016)]{chen2016xgboost}
Tianqi Chen and Carlos Guestrin.
\newblock Xgboost: A scalable tree boosting system.
\newblock In \emph{Proceedings of the 22nd acm sigkdd international conference on knowledge discovery and data mining}, pages 785--794, 2016.

\bibitem[Chiba et~al.(2022)Chiba, Suzuki, Taniai, Igarashi, Ushiku, Saito, and Ono]{chiba2022neural}
Naoya Chiba, Yuta Suzuki, Tatsunori Taniai, Ryo Igarashi, Yoshitaka Ushiku, Kotaro Saito, and Kanta Ono.
\newblock Neural structure fields with application to crystal structure autoencoders.
\newblock \emph{arXiv preprint arXiv:2212.13120}, 2022.

\bibitem[Court et~al.(2020)Court, Yildirim, Jain, and Cole]{court20203}
Callum~J Court, Batuhan Yildirim, Apoorv Jain, and Jacqueline~M Cole.
\newblock 3-d inorganic crystal structure generation and property prediction via representation learning.
\newblock \emph{Journal of Chemical Information and Modeling}, 60\penalty0 (10):\penalty0 4518--4535, 2020.

\bibitem[G{\'o}mez-Bombarelli et~al.(2018)G{\'o}mez-Bombarelli, Wei, Duvenaud, Hern{\'a}ndez-Lobato, S{\'a}nchez-Lengeling, Sheberla, Aguilera-Iparraguirre, Hirzel, Adams, and Aspuru-Guzik]{gomez2018}
Rafael G{\'o}mez-Bombarelli, Jennifer~N Wei, David Duvenaud, Jos{\'e}~Miguel Hern{\'a}ndez-Lobato, Benjam{\'\i}n S{\'a}nchez-Lengeling, Dennis Sheberla, Jorge Aguilera-Iparraguirre, Timothy~D Hirzel, Ryan~P Adams, and Al{\'a}n Aspuru-Guzik.
\newblock Automatic chemical design using a data-driven continuous representation of molecules.
\newblock \emph{ACS central science}, 4\penalty0 (2):\penalty0 268--276, 2018.

\bibitem[Griffiths and Hernández-Lobato(2020)]{griffiths_etal_2020}
Ryan-Rhys Griffiths and José~Miguel Hernández-Lobato.
\newblock Constrained bayesian optimization for automatic chemical design using variational autoencoders.
\newblock \emph{Chem. Sci.}, 11:\penalty0 577--586, 2020.
\newblock \doi{10.1039/C9SC04026A}.

\bibitem[Grosnit et~al.(2021)Grosnit, Tutunov, Maraval, Griffiths, Cowen-Rivers, Yang, Zhu, Lyu, Chen, Wang, Peters, and Bou-Ammar]{Grosnit2021HighDimensionalBO}
Antoine Grosnit, Rasul Tutunov, Alexandre~Max Maraval, Ryan-Rhys Griffiths, Alexander~Imani Cowen-Rivers, Lin Yang, Lin Zhu, Wenlong Lyu, Zhitang Chen, Jun Wang, Jan Peters, and Haitham Bou-Ammar.
\newblock High-dimensional bayesian optimisation with variational autoencoders and deep metric learning.
\newblock \emph{ArXiv}, abs/2106.03609, 2021.

\bibitem[Higgins et~al.(2016)Higgins, Matthey, Pal, Burgess, Glorot, Botvinick, Mohamed, and Lerchner]{higgins2016}
Irina Higgins, Lo{\"i}c Matthey, Arka Pal, Christopher~P. Burgess, Xavier Glorot, Matthew~M. Botvinick, Shakir Mohamed, and Alexander Lerchner.
\newblock beta-vae: Learning basic visual concepts with a constrained variational framework.
\newblock In \emph{International Conference on Learning Representations}, 2016.

\bibitem[Hoffmann et~al.(2019)Hoffmann, Maestrati, Sawada, Tang, Sellier, and Bengio]{hoffmann2019data}
Jordan Hoffmann, Louis Maestrati, Yoshihide Sawada, Jian Tang, Jean~Michel Sellier, and Yoshua Bengio.
\newblock Data-driven approach to encoding and decoding 3-d crystal structures.
\newblock \emph{arXiv preprint arXiv:1909.00949}, 2019.

\bibitem[Jain et~al.(2013)Jain, Ong, Hautier, Chen, Richards, Dacek, Cholia, Gunter, Skinner, Ceder, and Persson]{Jain2013}
Anubhav Jain, Shyue~Ping Ong, Geoffroy Hautier, Wei Chen, William~Davidson Richards, Stephen Dacek, Shreyas Cholia, Dan Gunter, David Skinner, Gerbrand Ceder, and Kristin~A. Persson.
\newblock Commentary: the materials project: a materials genome approach to accelerating materials innovation.
\newblock \emph{APL Materials}, 1:\penalty0 011002, 2013.
\newblock \doi{10.1063/1.4812323}.
\newblock URL \url{https://doi.org/10.1063/1.4812323}.

\bibitem[Kim et~al.(2020)Kim, Noh, Gu, Aspuru-Guzik, and Jung]{kim2020generative}
Sungwon Kim, Juhwan Noh, Geun~Ho Gu, Alan Aspuru-Guzik, and Yousung Jung.
\newblock Generative adversarial networks for crystal structure prediction.
\newblock \emph{ACS central science}, 6\penalty0 (8):\penalty0 1412--1420, 2020.

\bibitem[Kingma and Welling(2014)]{kingma_welling_2014}
Diederik~P. Kingma and Max Welling.
\newblock Auto-encoding variational bayes.
\newblock In Yoshua Bengio and Yann LeCun, editors, \emph{2nd International Conference on Learning Representations, {ICLR} 2014}, 2014.

\bibitem[Krenn et~al.(2020)Krenn, Häse, Nigam, Friederich, and Aspuru-Guzik]{krenn_2020}
Mario Krenn, Florian Häse, AkshatKumar Nigam, Pascal Friederich, and Alan Aspuru-Guzik.
\newblock Self-referencing embedded strings (selfies): A 100\% robust molecular string representation.
\newblock \emph{Machine Learning: Science and Technology}, 1\penalty0 (4):\penalty0 045024, oct 2020.
\newblock \doi{10.1088/2632-2153/aba947}.

\bibitem[Maus et~al.(2022)Maus, Jones, Moore, Kusner, Bradshaw, and Gardner]{maus2022local}
Natalie Maus, Haydn Jones, Juston Moore, Matt~J Kusner, John Bradshaw, and Jacob Gardner.
\newblock Local latent space bayesian optimization over structured inputs.
\newblock \emph{Advances in Neural Information Processing Systems}, 35:\penalty0 34505--34518, 2022.

\bibitem[Noh et~al.(2019)Noh, Kim, Stein, Sanchez-Lengeling, Gregoire, Aspuru-Guzik, and Jung]{nohinverse}
Juhwan Noh, Jaehoon Kim, Helge~S. Stein, Benjamin Sanchez-Lengeling, John~M. Gregoire, Alan Aspuru-Guzik, and Yousung Jung.
\newblock Inverse design of solid-state materials via a continuous representation.
\newblock \emph{Matter}, 1\penalty0 (5):\penalty0 1370--1384, 2019.
\newblock ISSN 2590-2385.
\newblock \doi{https://doi.org/10.1016/j.matt.2019.08.017}.
\newblock URL \url{https://www.sciencedirect.com/science/article/pii/S2590238519301754}.

\bibitem[Ong et~al.(2013)Ong, Richards, Jain, Hautier, Kocher, Cholia, Gunter, Chevrier, Persson, and Ceder]{pymatgen}
Shyue~Ping Ong, William~Davidson Richards, Anubhav Jain, Geoffroy Hautier, Michael Kocher, Shreyas Cholia, Dan Gunter, Vincent~L. Chevrier, Kristin~A. Persson, and Gerbrand Ceder.
\newblock Python materials genomics (pymatgen): A robust, open-source python library for materials analysis.
\newblock \emph{Computational Materials Science}, 68:\penalty0 314--319, 2013.
\newblock ISSN 0927-0256.
\newblock \doi{https://doi.org/10.1016/j.commatsci.2012.10.028}.
\newblock URL \url{https://www.sciencedirect.com/science/article/pii/S0927025612006295}.

\bibitem[Ren et~al.(2020)Ren, Noh, Tian, Oviedo, Xing, Liang, Aberle, Liu, Li, Jayavelu, et~al.]{ren2020inverse}
Zekun Ren, Juhwan Noh, Siyu Tian, Felipe Oviedo, Guangzong Xing, Qiaohao Liang, Armin Aberle, Yi~Liu, Qianxiao Li, Senthilnath Jayavelu, et~al.
\newblock Inverse design of crystals using generalized invertible crystallographic representation.
\newblock \emph{arXiv preprint arXiv:2005.07609}, 3\penalty0 (6):\penalty0 7, 2020.

\bibitem[Ren et~al.(2022)Ren, Tian, Noh, Oviedo, Xing, Li, Liang, Zhu, Aberle, Sun, et~al.]{ren2022invertible}
Zekun Ren, Siyu Isaac~Parker Tian, Juhwan Noh, Felipe Oviedo, Guangzong Xing, Jiali Li, Qiaohao Liang, Ruiming Zhu, Armin~G Aberle, Shijing Sun, et~al.
\newblock An invertible crystallographic representation for general inverse design of inorganic crystals with targeted properties.
\newblock \emph{Matter}, 5\penalty0 (1):\penalty0 314--335, 2022.

\bibitem[Tripp et~al.(2020)Tripp, Daxberger, and Hernández-Lobato]{tripp_etal_2020}
Austin Tripp, Erik Daxberger, and José Hernández-Lobato.
\newblock Sample-efficient optimization in the latent space of deep generative models via weighted retraining.
\newblock In \emph{Advances in Neural Information Processing Systems}, 06 2020.

\bibitem[Weininger(1988)]{weininger_1988}
David Weininger.
\newblock Smiles, a chemical language and information system. 1. introduction to methodology and encoding rules.
\newblock \emph{Journal of Chemical Information and Computer Sciences}, 28\penalty0 (1):\penalty0 31--36, 1988.
\newblock \doi{10.1021/ci00057a005}.

\bibitem[Xie et~al.(2021)Xie, Fu, Ganea, Barzilay, and Jaakkola]{xie2021crystal}
Tian Xie, Xiang Fu, Octavian-Eugen Ganea, Regina Barzilay, and Tommi Jaakkola.
\newblock Crystal diffusion variational autoencoder for periodic material generation.
\newblock \emph{arXiv preprint arXiv:2110.06197}, 2021.

\bibitem[Zeni et~al.(2023)Zeni, Pinsler, Z{\"u}gner, Fowler, Horton, Fu, Shysheya, Crabb{\'e}, Sun, Smith, et~al.]{zeni2023mattergen}
Claudio Zeni, Robert Pinsler, Daniel Z{\"u}gner, Andrew Fowler, Matthew Horton, Xiang Fu, Sasha Shysheya, Jonathan Crabb{\'e}, Lixin Sun, Jake Smith, et~al.
\newblock Mattergen: a generative model for inorganic materials design.
\newblock \emph{arXiv preprint arXiv:2312.03687}, 2023.

\bibitem[Zhao et~al.(2021)Zhao, Al-Fahdi, Hu, Siriwardane, Song, Nasiri, and Hu]{zhao2021high}
Yong Zhao, Mohammed Al-Fahdi, Ming Hu, Edirisuriya~MD Siriwardane, Yuqi Song, Alireza Nasiri, and Jianjun Hu.
\newblock High-throughput discovery of novel cubic crystal materials using deep generative neural networks.
\newblock \emph{Advanced Science}, 8\penalty0 (20):\penalty0 2100566, 2021.

\bibitem[Zhao et~al.(2023)Zhao, Siriwardane, Wu, Fu, Al-Fahdi, Hu, and Hu]{zhao2023physics}
Yong Zhao, Edirisuriya M~Dilanga Siriwardane, Zhenyao Wu, Nihang Fu, Mohammed Al-Fahdi, Ming Hu, and Jianjun Hu.
\newblock Physics guided deep learning for generative design of crystal materials with symmetry constraints.
\newblock \emph{npj Computational Materials}, 9\penalty0 (1):\penalty0 38, 2023.

\end{thebibliography}

\appendix

\section{Appendix / supplemental material}

\subsection{LCA-LSBO}\label{sec:app1}

Along with the problem of limited labeled data obtained from BB function evaluations to update the model discussed in \S\ref{sec:lca-lsbo}, LCA-LSBO addresses another issue in current LSBO settings, which is the \emph{latent consistency/inconsistency.} A point in the latent space is considered to be latent consistent when its location remains unchanged after being processed through the decoder and then the encoder of the VAE, which means latent consistency is achieved when $\bm{z} \approx f^{\text{enc}}_\phi(f^{\text{dec}}_{\theta}(\bm{\hat{z}}))$.Therefore, the \emph{reconstruction} objective discussed in \S\ref{sec:lca-lsbo} aims to improve the latent consistency in the latent space. Latent inconsistencies cause LSBO algorithms not to operate as intended. Namely, as the GP is updated with the queried point $\bm{\hat{z}}$ and the VAE is updated with the $f^{\text{dec}}_{\theta}(\bm{\hat{z}})$, due to the inconsistencies between $\bm{\hat{z}}$ and $f^{\text{enc}}_\phi(f^{\text{dec}}_{\theta}(\bm{\hat{z}}))$, BO loses the location information of the previous iterations, including promising regions that worth further exploitation. Furthermore, the discrepancy between $\bm{\hat{z}}$ and $f^{\text{enc}}_\phi(f^{\text{dec}}_{\theta}(\bm{\hat{z}}))$ grows as the density of the sampling region of $\bm{\hat{z}}$ decreases, leading to noticeable latent inconsistencies particularly during the exploration phase of the LSBO. On the other hand, while the retraining aims to enhance the VAE's ability to generate target instances by incorporating newly queried data, the limited number of these instances often fails to produce meaningful updates to the VAE. Given the high cost of BB function evaluations, it is impractical to expect a sufficient volume of BB function queries to gather enough new instances for significant VAE updates.

In LCA-LSBO, augmented instances are generated from $p^{\text{ref}}$, which is referred to as \emph{latent reference distribution} with mean $\bm{\mu}^{\text{ref}}$ and standard deviation $\sigma^{\text{ref}}$. This distribution focuses on the augmentation of synthetic latent variables near regions of interest by setting $\sigma^{\text{ref}} < 1$, allowing LCA-LSBO to target specific areas for updates. Once augmented instances are drawn, the VAE model is retrained using an additional term reconstruction term for latent data augmentations in its objective function. This term is referred to as Latent Consistency Loss (LCL). The LCL penalizes deviations between latent variables $\bm{\hat{z}}$ and $f^{\text{enc}}_\phi(f^{\text{dec}}_{\theta}(\bm{\hat{z}}))$. The LCL aims to mitigate the negative impacts of the latent inconsistencies between $\bm{\hat{z}}$ and $f^{\text{enc}}_\phi(f^{\text{dec}}_{\theta}(\bm{\hat{z}}))$ to LSBO performance. 

By incorporating label-free data augmentations in the latent space through LCL, LCA-LSBO simultaneously addresses latent inconsistency and increases the number of instances available for retraining, focusing on the region of interest. The VAE variant utilizing LCL is known as Latent Consistent-Aware VAE (LCA-VAE), with its objective function defined as:
\begin{equation}
\label{eq:objFnc-LRA-VAE}
\mathcal{L}^{\text{LCA}}_{\text{VAE}}(\phi, \theta) = \mathcal{L}_{\text{VAE}}(\phi, \theta) - \gamma \mathbb{E}_{\bm{\hat{z}} \sim p_{\text{ref}}(\bm{\hat{z}})}\left[ \text{LCL}(\bm{\hat{z}}) \right],
\end{equation}
where $\text{LCL}(\bm{\hat{z}}) = ||\bm{\hat{z}} - f^{\text{enc}}_\phi(f^{\text{dec}}_{\theta}(\bm{\hat{z}}))||^2$ and the $\gamma$ is the hyperparameter of the LCL term, controlling its weight. By optimizing this objective function, LCA-VAE shapes the latent space by integrating the data augmentations into the retraining process, enhancing exploration capabilities, and improving sample efficiency. Therefore, in Crystal-LCA-LSBO, Lattice-VAE, Coordinate-VAE, and Element-VAE models are retrained using the LCL term in their objective function,
\begin{equation}
\label{eq:objFnc-LRA-VAE}
\mathcal{L}^{\text{VAE}^{\text{RT}}}_{\text{Crystal-LSBO}}(\phi, \theta) = \mathcal{L}^{\text{VAE}}_{\text{Crystal-LSBO}}(\phi, \theta) - \gamma \mathbb{E}_{\bm{\hat{z}} \sim p_{\text{ref}}(\bm{\hat{z}})}\left[ \text{LCL}(\bm{\hat{z}}) \right],
\end{equation}
where $\mathcal{L}^{\text{VAE}^{\text{RT}}}_{\text{Crystal-LSBO}}(\phi, \theta)$ denotes the updated objective function of Lattive-VAE, Coordinate-VAE, and Element-VAE models during retraining.

\subsection{Space Group 1 and Its Implications for VAE Training}\label{sec:spacegroup1}


Space groups are a way to categorize the patterns and symmetry seen in crystal structures. These patterns are formed by symmetry operations such as translations, rotations, reflections, and inversions that are applied to the arrangement of chemical elements. There are 230 different space groups, each representing a unique set of these operations, which create the repeating patterns in crystals \cite{brock2016international}. Space group 1 is the simplest of all these groups. It involves only translations, which means the pattern repeats itself periodically without additional symmetry operations such as rotations or reflections. This group imposes the fewest geometric constraints: the angles and lengths of the edges of the basic unit of the crystal can vary freely. Thanks to these properties, all crystals can be represented using space group 1; that is, representations from other space groups can be converted to space group 1, albeit in a simplified form without the complex symmetry operations found in other groups.

We chose to use space group 1 in our study because it's the simplest representation and offers the most flexibility. Since crystals in this group can take any shape or size, we can work with a wide range of structures. This is particularly useful when training our VAEs because it helps the model learn from a diverse set of crystal structures with their simplified representations. This diversity improves the model’s ability to generate new structures or to better understand ones it hasn't seen before. On the other hand, by using space group 1 representations, we inherently miss out on incorporating symmetry information into our framework. 



\subsection{Training Details of VAE Models}\label{sec:app2}

We utilized the PyTorch library to train all our models. For the Lattice-VAE, given the simplicity of its inputs—angles and cell lengths—we used linear layers in both the encoder and decoder. In contrast, the Coordinate-VAE, Element-VAE, and Combined-VAE models featured Convolutional Neural Network (CNN) based layers in encoders and decoders, which are followed by batch normalization, and leaky ReLU operations. Each VAE's property prediction model leveraged latent representations to estimate formation energy and band gap properties, using linear layers with ReLU activations. We calculated the MSE for these predictions and conducted joint training of this model with the VAE, utilizing the RMSProp optimizer throughout the training process. 

In our experiments, we utilized a computing environment equipped with two Intel® Xeon® Gold 6130 CPUs with a total of 32 cores, which are part of internal cluster machine. In this setup, the training of models requires approximately 3 hours, each LSBO experiment is completed in under one hour per seed, and random generation experiments conclude in less than 10 minutes per seed. 

\subsection{Hyperparameter Optimization for LCA-LSBO}\label{sec:app3}

In the Crystal-LCA-LSBO framework, we need to optimize hyperparameters: $p^{\rm ref}$ standard deviation $\sigma^{\text{ref}}$ which is shared across $p^{\rm ref}_{\rm Lattice}$, $p^{\rm ref}_{\rm Coordinate}$, and $p^{\rm ref}_{\rm Element}$, the threshold for the region of interest $\tau$, and the weight of the LCL term $\gamma$. Optimization was conducted within the surrogate-LSBO experimental setup detailed in \S\ref{sec:lsbo-model-selection}, where electronegativity was the target property. Using the selected model in \S \ref{sec:lsbo-model-selection}, we run several LSBO experiments with $\sigma^{\text{ref}}$ values ranging from $\sigma^{\text{ref}} \in [0.3, 0.5, 0.7, 1]$, $\gamma \in [0.1, 1, 2.5, 5, 7.5, 10]$, and $\tau$ set to percentile thresholds $P_{50}$ (median), $P_{75}$, and $P_{100}$ (maximum), based on electronegativity values in our dataset. Each experiment is repeated with 200 iterations and 10 different seeds. The optimal combination of hyperparameters was found to be $\sigma^{\text{ref}}=0.3$, $\gamma=2.5$, and $\tau=P_{100}$. The sample size of augmentations, $N^*$, is set to match the batch size used in the model training/retraining.

\end{document}